\documentclass[journal]{IEEEtran}
\usepackage[ printonlyused,withpage]{acronym}
\usepackage{multirow,tabularx}
\usepackage{colortbl}
\usepackage{xspace}
\usepackage{hyperref}
\usepackage{graphicx}
\usepackage{cite}
\usepackage{amssymb,amsfonts}
\usepackage[fleqn]{amsmath}
\usepackage{algorithmic}
\usepackage{textcomp}
\usepackage{xcolor}
\usepackage[ruled,vlined]{algorithm2e}
\usepackage{multirow}
\usepackage{booktabs}
\usepackage{subcaption}
\usepackage{hyperref}
\usepackage{todonotes}
\usepackage{color,soul}
\usepackage{xcolor}
\usepackage{svg}
\usepackage{cleveref}
\usepackage{nicematrix}
\usepackage{cleveref} 
\acrodef{avs}[AV]{Autonomous Vehicles}
\acrodef{lcf}[LCF]{Local Contextual Features}
\acrodef{gf}[GF]{Global Features}
\acrodef{bev}[BEV]{Bird-Eye-View}
\acrodef{lstm}[LSTM-VAE]{A Long Short-Term Memory Variational Autoencoder}
\acrodef{clstm}[C-LSTM]{Convolutional Long Short-Term Memory Network}
\acrodef{pti}[PTINet]{Pedestrain Trajectory and Intention prediction Network }
\acrodef{ade}[ADE]{Average Displacement Error }
\acrodef{fde}[FDE]{Final Displacement Error }
\acrodef{vae}[VAE]{Variational Autoencoders}
\ifCLASSINFOpdf
\else
\fi
\usepackage{ifthen}

\newboolean{commenttingcomands}
\setboolean{commenttingcomands}{true}

\newcommand{\add}[2]{%
  \ifthenelse{\boolean{commenttingcomands}}%
  {%
    \textbf{\textcolor{green}{#1[ADD]:#2}}%
  }%
  {%
  }%
}

\newcommand{\remove}[2]{%
  \ifthenelse{\boolean{commenttingcomands}}%
  {%
    \textbf{\textcolor{red}{#1[REMOVE]:#2}}%
  }%
  {%
    #2%
  }%
}

\newcommand{\replace}[3]{%
  \ifthenelse{\boolean{commenttingcomands}}%
  {%
    \textbf{\textcolor{red}{#1:[REPLACE]:#2}\textcolor{green}{[WITH]:#3}}%
  }%
  {%
    #1%
  }%
}

\newcommand{\comment}[2]{%
  \ifthenelse{\boolean{commenttingcomands}}%
  {%
    \textbf{\textcolor{blue}{#1:[COMMENT]:#2}}%
  }%
  {%
  }%
}

\newcommand{\commenton}[3]{%
  \ifthenelse{\boolean{commenttingcomands}}%
  {%
    \textbf{\textcolor{blue}{#1:[TEXT]:#2[COMMENT]:#3}}%
  }%
  {%
    #2%
  }%
}

\begin{document}
\title{ Context-aware Multi-task Learning for Pedestrian  Intent and  Trajectory Prediction}
% \title{Multi-Task Learning for Context-Aware Pedestrian Behavior Prediction in Autonomous Driving Systems}

\author{\IEEEauthorblockN{Farzeen Munir\textsuperscript{1,2}, Tomasz Piotr Kucner \textsuperscript{1,2}}\\
\IEEEauthorblockA{\textsuperscript{1}Department of Electrical Engineering and Automation,
Aalto University,
 Finland\\
}
\IEEEauthorblockA{\textsuperscript{2}Finnish Center for Artificial Intelligence,
Finland\\
Email: {(farzeen.munir,  tomasz.kucner)@aalto.fi}
}
}


\markboth{Journal of \LaTeX\ Class Files,~Vol.~14, No.~8, August~2015}%
{Shell \MakeLowercase{\textit{et al.}}: Bare Demo of IEEEtran.cls for IEEE Journals}



\maketitle

\begin{abstract}

The advancement of socially-aware autonomous vehicles hinges on  precise modeling of human behavior. Within this broad paradigm, the specific challenge lies in accurately predicting pedestrian's trajectory and intention. Traditional methodologies have leaned heavily on historical trajectory data, frequently overlooking vital contextual cues such as pedestrian-specific traits and environmental factors. Furthermore, there's a notable knowledge gap as trajectory and intention prediction have largely been approached as separate problems, despite their mutual dependence. To bridge this gap,  we introduce PTINet (Pedestrian Trajectory and Intention Prediction Network), which jointly learns the trajectory and intention prediction by combining past trajectory observations, local contextual features (individual pedestrian behaviors), and global features (signs, markings etc.). The efficacy of our approach is evaluated on widely used public datasets: JAAD and PIE, where it has demonstrated superior performance over existing state-of-the-art models in trajectory and intention prediction. The results from our experiments and ablation studies robustly validate PTINet's effectiveness in jointly exploring intention and trajectory prediction for pedestrian behaviour modelling. The experimental evaluation indicates the advantage of using global and local contextual features for pedestrian trajectory and intention prediction. The effectiveness of PTINet in predicting pedestrian behavior paves the way for the development of automated systems capable of seamlessly interacting with pedestrians in urban settings.
% PTINet could potentially aids in developing automated systems to prevent pedestrian accidents, enhancing road safety by enabling vehicles to anticipate and respond to pedestrian behavior. 
Our source code is available at \url{https://github.com/aalto-mobile-robotics-group/PTINet.git}

% Modeling human behavior is critical for the development of socially-aware autonomous vehicles. 
% In this context, pedestrian behavior modeling is formulated as both trajectory and intention prediction. 
% Traditional approaches have primarily relied on historical trajectory data, often neglecting critical contextual information such as pedestrian-specific features and environmental attributes. Moreover, despite their inherent interdependence, most prior research has treated trajectory and intention prediction as distinct tasks. To bridge this gap,  we introduce PTINet (Pedestrian Trajectory and Intention Prediction Network), which jointly learns the trajectory and intention prediction by combining past trajectories, local context features, and global features. The local context features focus on individual pedestrian behaviors, whereas the global features encompass environmental information relevant to the pedestrian from the perspective of the ego-vehicle. The efficacy of our approach is evaluated on widely used public datasets: JAAD and PIE. Our results and ablation studies indicate that PTINet surpasses existing state-of-the-art approaches in trajectory and intention prediction. Our source code is available at \url{https://github.com/aalto-mobile-robotics-group/PTINet.git}
\end{abstract}


% \fi
\IEEEpeerreviewmaketitle

\section{Introduction}
\Acfi{avs} has evolved swiftly in recent years, with safety being paramount \cite{maurer2016autonomous}. A critical step towards safety involves accurately predicting pedestrian behaviour. This capability enables autonomous vehicles to identify and avoid potential collisions. For instance, the inability to anticipate a pedestrian's intention to cross the road leaves an autonomous vehicle with no choice but to initiate braking only when the pedestrian appears on the road. This limits the reaction time and significantly increases the risk of not stopping in time, potentially leading to accidents. Therefore, predicting pedestrian behavior efficiently and accurately is a crucial task for safe, safe human-\ac{avs} interactions.
% Therefore, estimating pedestrian trajectories and intentions efficiently and accurately is a crucial task for safe human-\ac{avs} interactions.
\par
Predicting pedestrian behavior presents a significant challenge due to the lack of access to their complete internal state, necessitating the use of external cues. Predicting pedestrian behaviors is crucial for safety, especially when autonomous vehicles navigate in shared urban spaces \cite{fang2018pedestrian}. The pedestrian's behavior is affected by two factors. The primary factor is the pedestrian's historical trajectory, which encapsulates their latent intent. The secondary factor concerns the environmental context, delineated by accessible and restricted areas \cite{sharma2022pedestrian}. A proficient model for predicting pedestrian behaviour necessitates the integration of these two critical factors. In research methodologies, modeling pedestrian behaviors involves two approaches \cite{rudenko2020human,herman2021pedestrian,kotseruba2021benchmark} that are categorized into i) \textbf{Intention Prediction} and ii) \textbf{Trajectory Prediction}. In this work, we poised the research question that predicting the former without considering the latter often limits understanding the pedestrian behavior in context to \ac{avs}.
\par
Intention prediction involves anticipating the pedestrian's next move. In the case of urban settings, intention prediction has been mostly studied in literature by modeling factors, such as predicting intentions by analyzing historical data on position and contextual features (i.e., gait\cite{minguez2018pedestrian}, activity \cite{lorenzo2020rnn}, and gestures\cite{gesnouin2022analysis}). While these elements are instrumental in deciphering pedestrian intentions, they predominantly model intentions from the individual's perspective without adequately accounting for environmental influences. In the literature, environmental factors and local attributes are also studied in addition to the above pedestrian-centric factors to model pedestrian intention prediction \cite{crosato2023social,millard2018pedestrians}. The core research challenge lies in integrating contextual and environmental factors into a cohesive framework for predicting pedestrian intentions, aiming to enhance behavior prediction for autonomous vehicles, even though visual cues have already proven effective in anticipation.

% \Acfi{avs} are revolutionizing transportation, becoming more prevalent in cities worldwide \cite{millard2018pedestrians}. These vehicles strive to provide safe, comfortable, and autonomous travel options. However, as \ac{avs} become more common, their integration with human road users becomes crucial. This integration poses questions regarding the \ac{avs}  capability to effectively comprehend and engage with human behaviors, potentially complicating traffic situations and elevating safety concerns \cite{maurer2016autonomous}.
% % Companies aim to offer safe, comfortable, and hands-free travel experiences. Yet, as their presence increases, they must coexist with human road users. This coexistence raises concerns about their ability to smoothly understand and interact with humans, which could lead to complex traffic scenarios and safety risks \cite{maurer2016autonomous}. 
% Pedestrian, being the most prominent entity of the traffic environment, needs smooth and reliable interaction with the \ac{avs} to develop a safe, dynamic environment for both parties. In response to design a safe environment, researchers are focusing on developing socially aware \ac{avs} tailored for safe and effective interaction with pedestrians \cite{crosato2023social,nishihori2020affects}. \ac{avs} need to understand human behavior and optimize their actions based on their interactions with other road users. Achieving this requires \ac{avs} to predict and adapt to the behaviors of nearby road users and making well-informed decisions\cite{ajzen2011theory}. 
\par 
\par

\begin{figure*}[t]
      \centering
      \includegraphics[width=15.5cm, keepaspectratio]{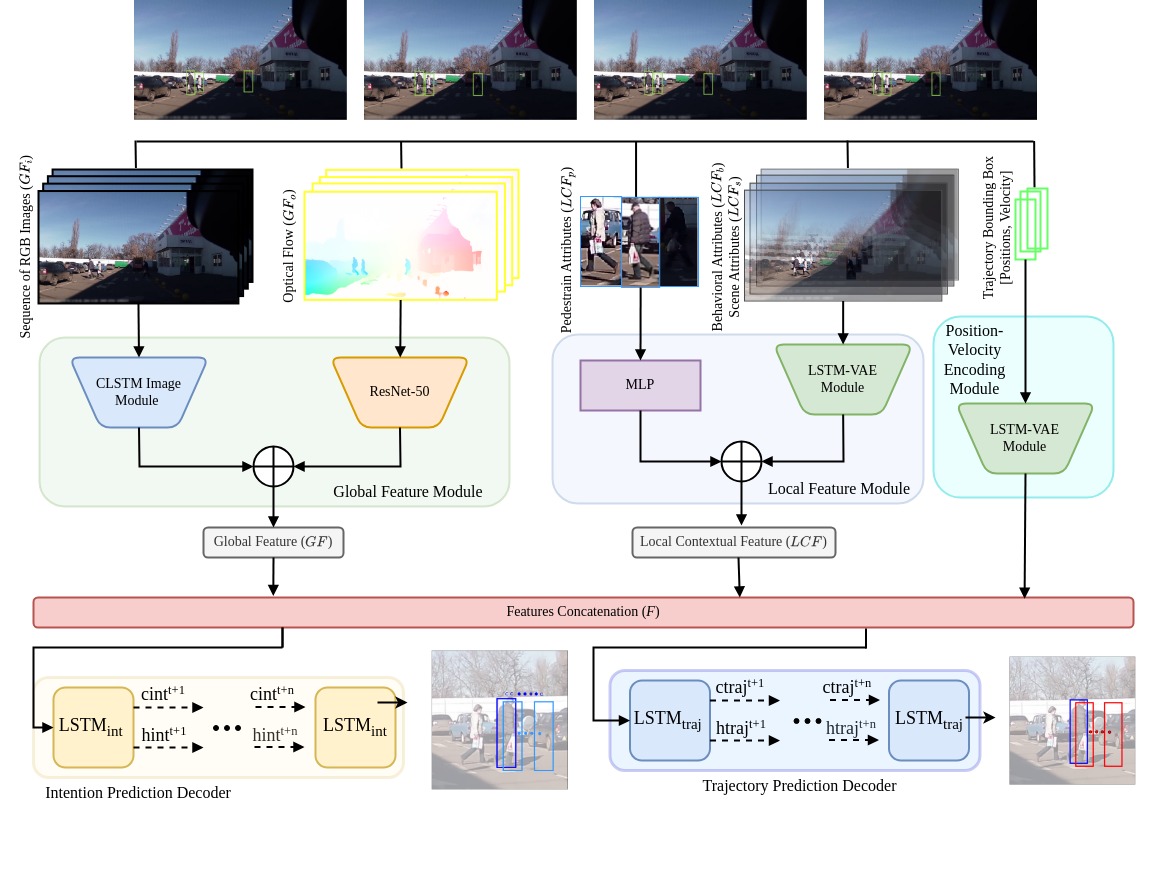}
      \caption{The figure illustrates a context-aware multi-task learning framework for the prediction of pedestrian trajectories and intentions. The architecture comprises a Global Feature Module, which processes image and optical flow data utilizing a \ac{clstm} module and Resnet50, respectively. Concurrently, the Local Contextual Module takes in local contextual features and employs a combination of MLP and LSTM-VAE blocks for feature extraction. The Position-Velocity Encoding Module encodes past pedestrian trajectories. The outputs from these distinct modules are concatenated and fed into separate trajectory and intention decoders, facilitating subsequent predictions}
      \label{framework}
\end{figure*}

% Since intention prediction is an interlinked problem with trajectory prediction, formulating the former without incorporating the latter \commenton{TK}{may not fully describe safe human AV interactions}{There is a bit of logical jump here. You say taht we need to integarte both and then you jump to the problem of safety. There is need to say how the integration will result in mechanisms that directly afect the safety.}.
% Trajectory prediction estimates the probable path a pedestrian is likely to follow \cite{rudenko2020human}. 
Since intention prediction is interlinked with trajectory prediction, formulating the former without incorporating the latter would not fully capture pedestrian's behavior, potentially resulting in unsafe human-\ac{avs} interactions. Integrating both pedestrian's intention and trajectory predictions enhances the ability of \ac{avs} systems to anticipate pedestrian movements more accurately, thereby directly improving safety mechanisms and reducing the likelihood of accidents.
In existing studies \cite{salzmann2020trajectron++,mohamed2022social,kothari2021human}, pedestrian trajectory prediction often relies only on past movements, ignoring its interdependence with intention prediction. Moreover, these studies usually overlook important contextual and environmental factors that are key to understanding human behavior. Recent studies \cite{gupta2018social,kothari2021human,yuan2021agentformer,rudenko2020human} have developed robust algorithms for human trajectory prediction, focusing on pedestrian interactions with surroundings and past trajectory data. Others have incorporated scene information through scene graphs \cite{salzmann2020trajectron++}, obstacle maps \cite{zhang2023obstacle}, and heat maps \cite{mangalam2021goals} to predict feasible trajectories. However, they are deficient in accurately predicting pedestrian behavior because they don't consider specific \ac{avs} features, pedestrian attributes, or traffic conditions. This limitation is particularly problematic in dynamic urban settings where various factors can cause pedestrian behavior to unpredictably change.
% Current intention prediction methods primarily concentrate on determining whether a pedestrian will cross the road. However, in our approach, we also predict the future trajectory, providing a more comprehensive understanding of pedestrian behavior to \ac{avs}.
\par
In this work, we introduces a comprehensive framework for \acfi{pti} that considers pedestrian past trajectory, \acfi{lcf}, and  \acfi{gf} to predict both trajectory and intention simultaneously. Our proposed framework (\ac{pti}) integrates past trajectories and visual data gathered by the ego vehicle's field-of-view camera, contrary to bird-eye-view perspective of data. The visual data is included in the model as \ac{lcf} and comprehensive \ac{gf}. The \ac{lcf}  capture attributes specific to pedestrians, including their behavior and the surrounding scene characteristics.
% The local contextual features encapsulate attributes specific to pedestrians, encompassing behavioral, pedestrian, and scene attributes. 
These features are represented as vectors, encompassing both pedestrian attributes like age and gender and their behaviors, such as gestures, gaze direction, movement, and nodding. Additionally, they include traffic-related information, for instance, pedestrian crossings, road types, traffic signs, number of lanes, and traffic signals. \ac{lcf} enable the model to understand and represent pedestrian behavior, capturing their immediate interactions, which are essential for accurate trajectory and intention prediction.
The \ac{gf} which consists of image data and optical flow derived from consecutive frames, are integrated into the model. Introducing image and optical flow data is particularly advantageous as it endows the framework with a more comprehensive understanding of the environment. Image data offers rich visual information, while optical flow enables the model to discern the temporal evolution of visual cues. The synergistic integration of past trajectories, \ac{gf} and \ac{lcf} is instrumental in discerning complex spatial-temporal patterns, ultimately enhancing the robustness and accuracy of intention and trajectory predictions within the proposed multi-task framework.

% This integration enhances the model's capability to understand the dynamics and context of the environment.

% Local contextual features empower the model to grasp pedestrians' immediate interactions, which is key for accurate short-term trajectory and intention forecasting.
% . This combination of global and local features ensures that predictions are more precise, context-sensitive, and safer across various environments. 
% This fusion of \ac{gf} and \ac{lcf} ensures that predictions are not only more precise and context-sensitive but also safer across a diverse range of environments.

% \item We have studied the pedestrian trajectory and intention prediction from the ego-vehicle perspective for developing socially acceptable autonomous vehicles.
 The main contributions of our work are:
\begin{enumerate}
\item We have developed a novel multi-task framework, PTINet, that integrates LCF, represented by pedestrian-specific attributes, and GF, embodied by image data and optical flow. This approach enables accurate prediction of pedestrian behaviors by analyzing complex spatial-temporal patterns.

\item To learn the spatial and temporal representation, we integrate \ac{clstm}, LSTM-VAE, and MLP in a unified encoder network, followed by an LSTM-based intention and trajectory prediction decoder.

\item Our experimental analysis and ablation studies show the efficacy of the proposed PTINet on widely used benchmark datasets, outperforming the state-of-the-art methods.

\end{enumerate}

% The addition of global context proves crucial in aiding the network by offering a broader understanding of the overall environment.

% Researchers view understanding pedestrian behaviors in \acp{av} as two interconnected challenges: predicting pedestrian intentions and forecasting their trajectories.
% Pedestrians, being the most vulnerable road users, necessitate particular focus from \acp{av} to understand the factors that influence their decision-making when crossing streets.

% To address these challenges, researchers are undertaking to advance socially aware \acp{av}. \acp{av}  should demonstrate human-like behavior and optimize their actions, considering interactions with other road users. To achieve this, \acp{av} needs the capability to predict the behavior of nearby users and make informed decisions. Special attention is required for pedestrians, as they are the most vulnerable road users. Understanding the factors influencing pedestrians' decisions on the road is important.
\section{Related Work}
% In AV research, predicting pedestrian behavior is increasingly important and typically falls into two categories. The first is intention prediction \cite{liu2020spatiotemporal,chen2021modeling,fang2018pedestrian,rasouli2017they,yau2021graph,lorenzo2020rnn}, which forecasts intention like crossing the road and is vital for immediate safety decisions. The second is trajectory prediction \cite{rasouli2019pie,alahi2016social,gupta2018social,salzmann2020trajectron++}, which estimates a pedestrian's future path and is useful for navigation. While the former provides essential qualitative data for immediate reactive measures, the latter offers quantitative positional data that can be integrated into the vehicle's navigation algorithms.
\subsection{Intention Prediction} 
Intention prediction is crucial for facilitating interactions between \ac{avs} and pedestrians \cite{liu2020spatiotemporal,chen2021modeling,fang2018pedestrian,rasouli2017they,yau2021graph,lorenzo2020rnn}, involving the prediction of pedestrians' future actions, such as the likelihood of crossing a road. This capability is vital for allowing \ac{avs} to make timely safety-related decisions.  Earlier work for intention prediction involved learning feature representation from a static driving scene \cite{kotseruba2016joint}, followed by improving feature representation by incorporating pose estimation of pedestrians for the intention prediction \cite{fang2018pedestrian}. Recent work includes the application of transformer networks \cite{zhang2023trep}, which are trained to extract temporal correlations from input features related to pedestrians in a video sequence. The networks simultaneously model pedestrian uncertainty and predict intentions. The most relevant state-of-the-art studies in pedestrian intent prediction closely aligned with our work include PIE-intent \cite{rasouli2019pie}, FF-STA \cite{yang2022predicting}, TAMformer\cite{osman2023tamformer}, PedFormer \cite{rasouli2023pedformer}, and BiPed \cite{rasouli2021bifold}. These studies are selected based on input modality, feature extraction methods, evaluation measures, and the benchmark datasets employed, that aligns with our pedestrian intention prediction settings. While serving as solid baseline approaches, these methods fall short in incorporating global context from the ego-vehicle perspective. For instance, FF-STA \cite{yang2022predicting}, PedFormer \cite{rasouli2023pedformer}, and BiPed \cite{rasouli2021bifold} segment the environment to model global context, potentially overlooking environmental dynamics. Conversely, PIE-intent \cite{rasouli2019pie} and TAMformer \cite{osman2023tamformer} heavily rely on local environmental context. In this work, we have selected these state-of-the-art methods that reflect quantitative and qualitative comparison with our proposed method in terms of modeling local contextual and global feature for predicting the pedestrian intentions. 
 The two state-of-the-art methods PedFormer \cite{rasouli2023pedformer} and BiPed \cite{rasouli2021bifold}, are closely aligned with our work that use LSTM-based network to predict both intention and trajectory using local images, past trajectories, and semantic segmentation maps. While these methodologies have significantly influenced our research, they come with certain constraints. For instance, both BiPed \cite{rasouli2021bifold} and PedFormer \cite{rasouli2023pedformer} integrate global data in the form of segmentation maps, which fails to capture subtle spatial-temporal dynamics and accurately model complex pedestrian behaviors. Our work distinguishes itself by integrating Local Context Features (\ac{lcf})  and comprehensive Global Features (\ac{gf}), utilizing a novel combination of images and optical flow to capture a more holistic understanding of the pedestrians and their environment. In contrast to PTINet, some other works for instance, PIE-intent employs a convolutional LSTM network to encode past visual data, combined with bounding box information to predict a pedestrian's intention \cite{rasouli2019pie}. Similarly, TAMformer uses features similar to PIE-intent but utilizes a transformer-based architecture for intention prediction \cite{osman2023tamformer}. FF-STA extracts pedestrian appearance and context features using two separate CNNs and pre-computed pose data \cite{yang2022predicting}.

\subsection{Trajectory Prediction}

Pedestrian trajectory prediction involves forecasting the future position of pedestrians based on their current and past locations, behaviors, and the surrounding environment. Trajectory prediction algorithms often rely on \acfi{bev} data and operate from a top-down perspective, which simplifies the calculation of relative distances between objects \cite{alahi2016social,sadeghian2019sophie,salzmann2020trajectron++}. For instance, Social LSTM uses a specialized pooling module to consider the influence of other agents \cite{alahi2016social}. Other approaches like  adversarial networks \cite{gupta2018social}  and MID algorithm \cite{mohamed2022social} also focus on modeling interactions. Trajectron++ incorporates semantic maps and dynamic constraints \cite{salzmann2020trajectron++}, while \cite{gu2022stochastic} employs a transformer-based model to capture temporal dependencies. Despite their advances, these methods generally rely on past trajectory data, limiting their accuracy in predicting complex human behavior, especially in the context of autonomous vehicles.
\par
In contrast to \ac{bev} approaches, some algorithms use a first-person perspective, adding complexity due to the ego vehicle's motion \cite{rasouli2019pie,cao2020using}. These methods mainly aim to predict pedestrian behavior by predicting trajectory. The trajectory prediction algorithm in this context uses diverse inputs like bounding boxes, ego-vehicle distance \cite{herman2021pedestrian}, and contextual information \cite{sui2021joint,yau2021graph}. Visual features and behavioral cues such as orientation and awareness level are also considered \cite{rasouli2017they,cao2020using,kooij2019context}. 
Despite integrating various features, the trajectory prediction algorithms show limited trajectory accuracy improvement on datasets like PIE \cite{rasouli2019pie,cao2020using}. 
In contrast to the methods mentioned above \cite{rasouli2019pie,cao2020using}, our approach incorporates \ac{lcf}, such as gesture, walk direction, and head nodding of pedestrians, as well as their attributes to enhance the prediction of future trajectories in the image plane. Additionally, we integrate \ac{gf} using image features and motion information from optical flow to improve overall scene understanding. 
\par
We propose that intention and trajectory prediction are interconnected aspects essential for accurately modeling pedestrian behavior from the perspective of an ego-vehicle. Addressing one without the other could result in an incomplete representation of pedestrian behavior, as both elements are crucial to understanding and predicting their actions in traffic scenarios.
\section{Methodology}
\subsection{Problem Formulation}
This study proposes a multi-task learning framework (\ac{pti}) to predict pedestrian trajectory and intention concurrently. In addition to pedestrian-centric features like key points, head orientations, and past trajectories, our approach extends its scope to include a more comprehensive set of features, especially \ac{gf} and \ac{lcf}, as shown by \cref{framework}. By incorporating these additional features, our goal is to capture the intricacies of human behavior more comprehensively, ultimately enhancing predictions for both trajectory and intentions. The formulation of our framework is outlined as follows.
\par
Given a video sequence $\mathcal{V}$ of an urban scenario, we define a sequence of observed video frames as $\mathcal{V} = \{f_1, f_2,...f_t\}$ where $t$ represents discrete time steps corresponding to individual image frames ($f_t$). Our approach aims to estimate the probability of a pedestrian's intention to cross the street, represented as $I \in [0,1]$, while concurrently predicting the pedestrian’s future trajectory. The trajectory of the pedestrian is characterized by a sequence of bounding boxes  $b_t=\{x_t,y_t, w_t, h_t\}$ where  $(x_t,y_t)$ is center coordinates, $w_t$ is the width, and $h_t$ is height in the $t$-th image frame. At a given time step
$t$, our framework predict the future trajectory, $	\Omega_{ft}=\{b_i^{t+1},...,b_i^{t+n}\}$ and the future intention to cross, $\Psi_{ft}=\{ I_i^{t+1},...,I_i^{t+n}\}$ for a pedestrian $i$ over a prediction horizon of $n$ time steps. This prediction is based on the pedestrian's past trajectory $\Omega_p$,  $LCF$, and $GF$ observed over a horizon $m$. The pedestrian past trajectory $\Omega_p$ encompasses both positions 
 $Pos_{p}=\{b_i^{t-m+1},...,b_i^{t}\}$ and velocity $Vel_{p}=\{bv_i^{t-m+1},...,bv_i^{t}\}$. The velocity $Vel_{p}$ at  $t$ is then estimated as the change in position from the previous frame $t-1$.
 \par
 The \ac{lcf} within this framework are categorized into pedestrian attributes, behavior, and scene attributes.
 \begin{figure*}[t]
      \centering
      \includegraphics[width=17cm, keepaspectratio]{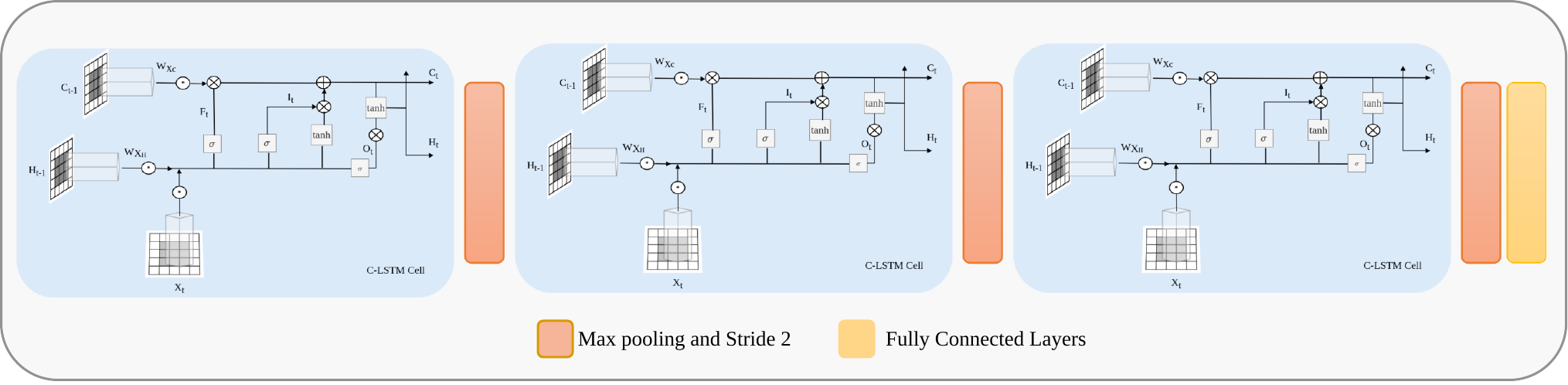}
      \caption{The \ac{clstm} module, designed to process input images and generate \ac{gf}. The detailed framework of the module comprises three blocks of \ac{clstm}, each followed by max pooling. The last block incorporates a max pooling layer followed by a fully connected layer.}
      \label{clstm}
\end{figure*}

 \begin{enumerate}
     \item Pedestrian Attributes $LCF_p$: These attributes are denoted as $LCF_p=\{ap_i, ....,ap_i \},$, where each $ap_i$ is a vector representing demographic aspects such as age and gender, and group size for each pedestrian.
     \item Behavior Attributes $LCF_b$: These attributes are articulated as $LCF_b=\{ab_i^{t-m+1}, ....,ab_i^{t} \}$, each $ab_i^t$ is a binary vector that consists of a range of non-verbal behavioral cues, such as looking, nodding, gesturing, and actions.
     \item Scene Attributes $LCF_s$: These attributes, represented as $LCF_s= \{as_i^{t-m+1}, \ldots, as_i^{t}\}$, comprise multidimensional vectors that intricately detail the environmental and infrastructural elements of the pedestrian's surroundings. Each vector  $as_i^t $  contains information on motion direction, number of lanes, traffic signs, pedestrian crossings, road types, and traffic signals.
 \end{enumerate}
% Lastly, the framework incorporates $GF$, which includes images and optical flow. 
% Optical flow is also exploited to compensate the dynam-
% ics of envoirment. 
Lastly, the framework incorporates $GF$, which include image data, denoted as $GF_{img}$, and optical flow, represented as $GF_o$. The image data is expressed as $GF_{img} =\{img^{t-m+1}, ....,img^{t} \}$ capturing the visual context from a series of frames, while the optical flow is detailed as $GF_o =\{of^{t-m+1}, ....,of^{t} \}$, quantifying the motion between these frames. The integration of optical flow is particularly important, as it enables the model to account for and adapt to the dynamic aspects of the environment. Optical flow offers an in-depth perspective on temporal variations and movements within the scene by analyzing motion patterns across sequential frames, which enhances the prediction of pedestrian behavior. 
\subsection{Architecture}

The framework depicted in \cref{framework} illustrates an integrated approach to predicting pedestrian trajectory and intention, using a combination of sequential image data, optical flow, and dynamic pedestrian attributes. The methodology adopts an encoder-decoder architecture, with each encoder module responsible for encoding the pedestrian past trajectory, \ac{lcf}, and \ac{gf} respectively.

\subsubsection{Position-Velocity Encoding Module}
\begin{figure}
      \centering
      \includegraphics[width=8cm, keepaspectratio]{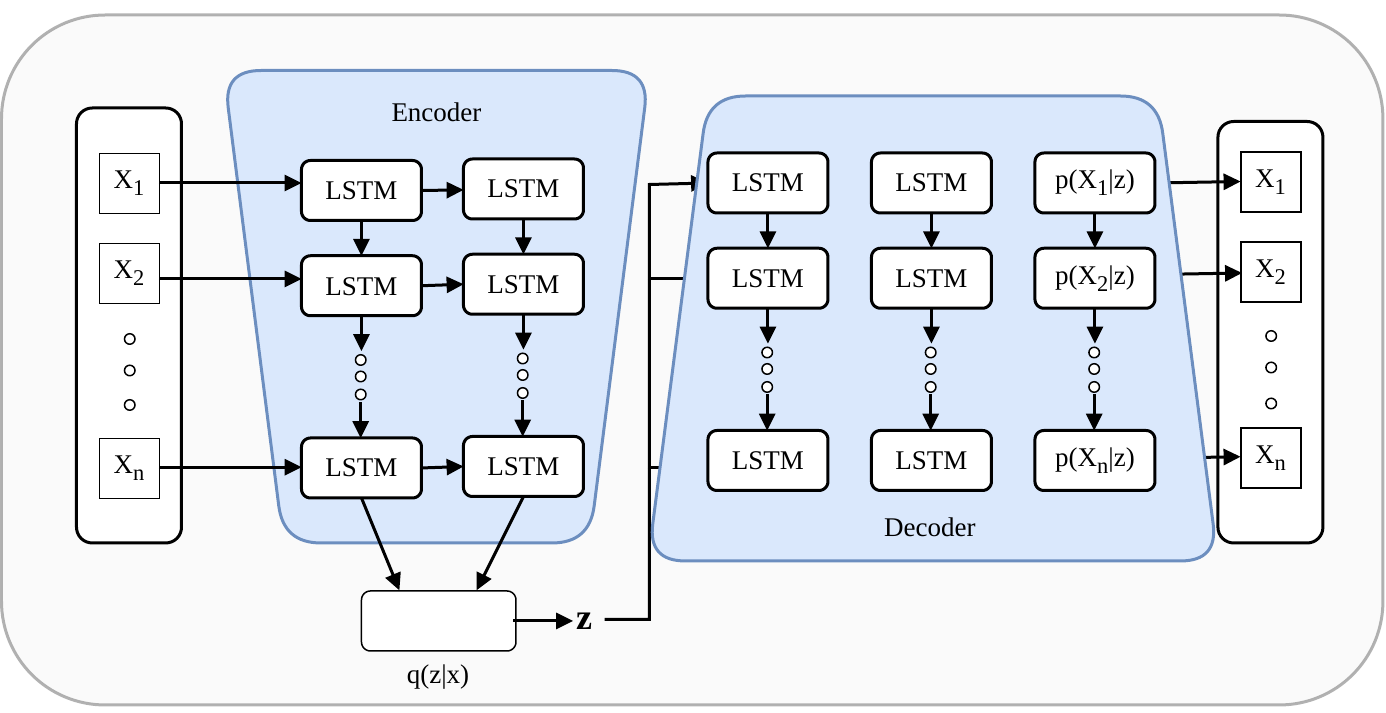}
      \caption{illustrates the architecture of the LSTM-VAE module employed in PTINet, which is utilized for learning the LCF and for capturing the temporal representation of past trajectories.}
      \label{lstm-vae}
\end{figure}

\acfi{lstm} block, as illustrated in  \cref{lstm-vae}, is employed for encoding the pedestrian trajectory consisting of pedestrian position $Pos_p$ and velocity $Vel_p$  \cite{hsu2017unsupervised}. \ac{lstm} for trajectory encoding is an optimal choice, as it effectively captures long-term dependencies and leverages the sequential data handling capability essential for maintaining temporal coherence in trajectory forecasting. Furthermore, the LSTM's ability to handle sequential data, combined with the generative modelling capabilities of \acfi{vae}, provides a comprehensive approach for accurately capturing the probabilistic nature of pedestrian movements and intentions. This block serves as a sequence-to-sequence autoencoder, encoding a sequence of input vectors into a latent space and then decoding from a sampled latent variable back to an input sequence. \ac{vae} are utilized to learn the generative process of pedestrian trajectories, while the \ac{lstm} block models the temporal relationships.
For the \ac{lstm} encoder, both the conditional distribution $p_\theta(x|z)$ and the approximate posterior distribution $q_\phi(z|x)$ are modeled as diagonal Gaussian distributions, as given by  \cref{eq7a} and \cref{eq7b}, respectively.
\begin{equation}
\label{eq7a}
\begin{aligned}
p_\theta(x|z)=\mathcal{N}(z:s_{\mu_z}(x;\theta),exp(s_{log\sigma_z^2 }(x;\theta))),
\end{aligned}
\end{equation}
\begin{equation}
\label{eq7b}
\begin{aligned}
q_\phi(z|x)=\mathcal{N}(z:g_{\mu_x}(z;\phi),exp(g_{log\sigma_x^2 }(z;\phi))), 
\end{aligned}
\end{equation}

Here, $s_{\mu_z}(x;\theta)$ and $g_{\mu_x}(z;\phi)$ represent mean and  $s_{log\sigma_z^2 }(x;\theta)$ and $ g_{log\sigma_x^2 }(z;\phi)$ represent the log variance, and are estimated by a neural network. The prior is set as a centered isotropic multivariate Gaussian $p_\theta(z)=\mathcal{N}(z;0,I)$ with 64 dimensions. The \ac{lstm} encoder consists of a two-layer LSTM with 512 hidden units to process the feature vector. Its outputs are merged and sent to a Gaussian layer to estimate the latent variable ($z$) mean and log variance. The reparameterization trick is applied to rewrite the latent variable as $z=s_{\mu_z}(x;\theta)+\sqrt{exp(s_{log\sigma _z^2}(x;\theta))}  \odot \varepsilon$
 where, $ \odot$ denotes element-wise multiplication, and $\varepsilon$ is sampled from $\mathcal{N}(z;0,I)$. The decoder of \ac{lstm} features a two-layer LSTM with 512 hidden units and takes the sampled latent variable to generate a sequence. Each generated output is then used as input for a Gaussian parameter layer, which predicts the mean and log variance for a single timestep of the input feature.

\subsubsection{Global Feature Module} 
This encoder integrates comprehensive global scene dynamics through image data and optical flow, capturing the dynamic changes and interactions that influence pedestrian movement. The image sequence undergoes processing via a \acfi{clstm} block, as depicted in \cref{clstm}. This block comprises three  \ac{clstm} cells \cite{shi2015convolutional}. Each \ac{clstm} cell is followed by a max-pooling layer, with the exception of the last cell, which is succeeded by a fully connected layer instead. Convolutional kernels for each layer have a size of 5x5 and a stride of 2x2, with 32 filters. This block is especially well-suited for image sequences, as it is designed to learn both spatial and temporal dependencies concurrently. \ac{clstm} cells maintain a continuously updated hidden state as they process the input sequence, enabling them to model non-linear temporal transitions effectively. Additionally, optical flow data are encoded using a ResNet-50 backbone, and the features extracted are amalgamated to form the \ac{gf}. 

  \subsubsection{Local Contextual Feature} It processes attributes directly related to pedestrians, such as demographic information, behavioral cues, and immediate environmental context, contributing to a holistic understanding of pedestrian behavior. Given the heterogeneous nature of the data, each attribute category is encoded distinctly to preserve its unique characteristics. Time-invariant pedestrian attributes are encoded utilizing a 64-layer Multi-Layer Perceptron (MLP) network optimized for static data representation. Conversely, pedestrian behavior attributes and scene attributes, which exhibit temporal variability, are processed using an \ac{lstm} block identical to the one previously described. This \ac{lstm} block models temporal dependencies and encodes them into a latent space. Such a space is probabilistically formulated to reflect the complex nature of pedestrian behavior and environmental factors, thus yielding a dense and informative representation of the active dynamics.
  \par

The encoded features $\mathbf{F}$, as shown in \cref{framework} from these modules (\ac{lcf} and \ac{gf})  are then fed into the decoder, which consists of the Trajectory Prediction Decoder and the Intention Prediction Decoder, which leverage the temporal and spatial context to predict future pedestrian trajectories and intentions.
\subsubsection*{Trajectory Prediction Decoder}
The trajectory decoder is designed to forecast pedestrian trajectories over a given timestep $n$. We opted for the LSTM because its inherent capacity to maintain long-term dependencies makes it ideally suited for the temporal precision required in trajectory forecasting.
An initial hidden state  $htraj^t$ is supplied to the trajectory decoder, $LSTM_{traj}$, which is the final concatenated feature vector $\mathbf{F}$ from the encoder module, as shown in \cref{framework}. This decoder takes the last observed position $Pos_b^t$ as its input and subsequently produces the next predicted position for the bounding box, expressed as $Pos_b^{t+1} = (x^{t+1}, y^{t+1}, w^{t+1}, h^{t+1})$
 The initial prediction is formulated through \cref{eq8}:
\begin{equation}
\label{eq8}
\begin{aligned}
 htraj^{t+1}=LSTM_{traj}(htraj^t,Pos_b^t,\mathcal{W}_{traj}) \\
\end{aligned}
\end{equation}

The predicted hidden state $htraj^{t+1}$ is then passed through a fully connected layer to calculate the output velocity, as described by \cref{eq9}:
\begin{equation}
\label{eq9}
\begin{aligned}
Pos_b^{t+1}=\mathcal{W}_ohtraj^{t+1}+bias_o \\
\end{aligned}
\end{equation}
Here, $\mathcal{W}_{traj}$ represents the weight matrix of the trajectory decoder, $\mathcal{W}_o$  is the weight matrix for the output layer, and $bias_o$  signifies its associated bias vector. Subsequent trajectory predictions are computed iteratively for a horizon $n$. In each iteration, the hidden state is updated, and the most recently predicted trajectory is provided as input to the decoder.
\subsubsection*{Pedestrian Intention Decoder}
Like the trajectory decoder, the intention decoder also employs LSTM network to process the encoded features from the previous modules, generating future intention predictions.
The intention decoder is initiated with a combined feature set $hint^{t}=\mathbf{F}$ as its initial hidden state. It also takes the last observed position of the bounding box, denoted as $Pos_b^{t} = (x^{t}, y^{t}, w^{t}, h^{t})$, as input. The decoder then outputs the subsequent predicted state of the pedestrian, $I^{t+1}$, as specified by \cref{eq10}.
\begin{equation}
\label{eq10}
\begin{aligned}
& hint^{t+1}=LSTM_{int}(hint^t,Pos_b^t,\mathcal{W}_{int}),\\
& I^{t+1}=\mathcal{W}_{oi}hint^{t+1}+bias_{oi}
\end{aligned}
\end{equation}
In this context, $LSTM_{int}$ represents the intention decoder, 
$\mathcal{W}_{int}$ is its weight matrix, $\mathcal{W}_{oi}$ is the weight matrix of the output layer, and $bias_{oi}$ is the associated bias vector. Subsequent pedestrian intentions for future timesteps $n$ are computed iteratively, with the hidden state being updated in each iteration. Finally, the output intentions are subjected to a softmax activation layer to calculate the probabilities associated with each potential outcome.
% The framework's core predictive capabilities are bifurcated into two decoding pathways: the Trajectory Prediction Decoder and the Intention Prediction Decoder (right). These decoders employ LSTM networks to process the encoded features from the previous modules, generating future trajectory paths and intention predictions. The LSTM networks' inherent capacity to maintain long-term dependencies makes them ideally suited for the temporal precision required in trajectory forecasting.
% These decoders are meticulously engineered to undertake the prediction of pedestrian trajectory and intention, embodying a critical function within the predictive modeling framework.

\subsection{Loss Funtions}
% The method is trained using a supervised approach. The Root Mean Squared Error (RMSE) serves as the loss function for trajectory prediction, quantifying the difference between the predicted bounding boxes, 
% $\hat{b}$, and the corresponding ground truth, $b$. This evaluation is performed for each of the $N$ training samples across a prediction horizon of $n$, as detailed in Equation \ref{eq11}.
% \begin{equation}
% \label{eq11}
% \begin{aligned}
% RMSE=\sqrt{\frac{1}{N}\sum_{i=1}^{N}\sum_{j=1}^{n}\left \|b_i^{t+j}-\hat{b}_i^{t+j}  \right \|}
% \end{aligned}
% \end{equation}

% For the task of intention prediction, the Binary Cross-Entropy (BCE) loss function is employed given by Equation \ref{eq12}. This choice is particularly well-suited for problems where the output can be categorized into one of two classes, such as predicting the intent of a pedestrian to either cross the street or not. The BCE loss function measures the divergence between the predicted probabilities, denoted as $\hat{I}$, and the actual ground truth labels, $I$, which are either 0 or 1.
% \begin{equation}
% \label{eq12}
% \begin{aligned}
% BCE=-\frac{1}{N}\sum_{i=1}^{N}(I_ilog(\hat{I}_i)+(1-I_i)log(1-\hat{I}_i))
% \end{aligned}
% \end{equation}
\begin{table*}[t]
    \centering
    \caption{Quantitative evaluation of the proposed method and state-of-the-art approaches on JAAD and PIE datasets for pedestrian trajectory prediction, focusing on \ac{ade} and \ac{fde} metrics across time horizons of 0.5s, 1s, and 1.5s}
    \label{tab:trajectory-prediction}
    \resizebox{15cm}{!}{%
    \begin{tabular}{lcccccccccccc}
        \toprule
        Methods & \multicolumn{6}{c}{JAAD } & \multicolumn{6}{c}{PIE }  \\
        \cmidrule(r){2-7}  \cmidrule(lr){8-13}  \\
          \textcolor{white}{Methods} & \multicolumn{3}{c}{ADE} & \multicolumn{3}{c}{FDE} & \multicolumn{3}{c}{ADE} & \multicolumn{3}{c}{FDE} \\
        \cmidrule(r){2-4} \cmidrule(lr){5-7} \cmidrule(lr){8-10} \cmidrule(l){11-13}
        & 0.5s & 1.0s & 1.5s & 0.5s & 1.0s & 1.5s & 0.5s & 1.0s & 1.5s & 0.5s & 1.0s & 1.5s \\
        \midrule
        Linear \cite{rasouli2019pie} & 233 & 857 & 2303 & - & - & 6111 & 123 & 477 & 1365 & - & - & 3983 \\
        LSTM \cite{rasouli2019pie} & 289 & 569 & 1558 & - & - & 5766 & 172 & 330 & 911 & - & - & 3352 \\
        B-LSTM \cite{rasouli2019pie} & 159 & 539 & 1535 & - & - & 5615 & 101 & 296 & 855 & - & - & 3259 \\ 
        PIE-intent \cite{rasouli2019pie} & 110 & 399 & 1248 & - & - & 4780 & 58 & 200 & 636 & - & - & 2477 \\
        Bi-Trap-D \cite{yao2021bitrap} & 93 & 378 & 1206 & - & - & 4565 & 41 & 161 & 511 & - & - & 1949 \\
        Bi-Trap-NP \cite{yao2021bitrap}  & 38 & 94 & 222 & - & - & 565 & 23 & 48 & 102 & - & - & 261 \\
        Bi-Trap-GMM \cite{yao2021bitrap} & 153 & 250 & 585 & - & - & 998 & 38 & 90 & 209 & - & - & 368 \\
        SGNet \cite{wang2022stepwise} & 82 & 328 & 1049 & - & - & 4076 & 34 & 133 & 442 & - & - & 1761 \\
        BiPed \cite{rasouli2021bifold} & - & 27.98 & - & - & 55.07 & - & - & 19.62 & - & - & 39.12& - \\
        PedFormer \cite{rasouli2023pedformer} & - & 24.56 & - & - & 48.82 & - & - & 15.27 & - & - & 32.79 & - \\
         \midrule
        % \textbf{Ours} & \textbf{12.50} & \textbf{25.60} & \textbf{44.42} & \textbf{24.05} & \textbf{57.58} & \textbf{102.21} & \textbf{4.99} & \textbf{10.41} & \textbf{18.23} & \textbf{9.96} & \textbf{25.31} & \textbf{52.20}\\
        \textbf{PTINet}   & \textbf{11.25} & \textbf{22.26} & \textbf{42.05} & \textbf{20.60} & \textbf{46.30} & \textbf{98.23} & \textbf{4.246} & \textbf{9.495} & \textbf{16.942} & \textbf{9.012} & \textbf{23.154} & \textbf{49.025} \\    
        % \textbf{OF}   & \textbf{19.2475} & \textbf{29.47} & \textbf{50.78} & \textbf{38.18} & \textbf{58.14} & \textbf{110.95} & \textbf{8.2772} & \textbf{17.57} & \textbf{26.68} & \textbf{16.44} & \textbf{33.89} & \textbf{61.87} \\
        \bottomrule
    \end{tabular}%
    }
\end{table*}

The proposed method loss includes two components: trajectory bounding box prediction loss ($\mathcal{L_{\text{traj}}}$), and intention prediction loss ($\mathcal{L_{\text{int}}}$). 
% The ($\mathcal{L_{\text{traj}}}$) loss is modeled by using LSTM-VAE in the proposed method.
The ($\mathcal{L_{\text{traj}}}$) is formulated as a combination of the reconstruction loss and the Kullback-Leibler (KL) divergence, which encourages the learned latent space to adhere to a predefined Gaussian distribution. Specifically, the reconstruction loss quantifies the discrepancy between the predicted bounding boxes and the ground truth, facilitating the model's ability to predict future states accurately. The KL divergence serves as a regularization term, ensuring that the distribution of the latent variables does not deviate significantly from the prior distribution. Mathematically, the trajectory bounding box prediction loss ($\mathcal{L_{\text{traj}}}$) is expressed as:
\begin{equation}
\label{traj}
\mathcal{L_{\text{traj}}} = \sum_{t=1}^{T} \left[ \beta \cdot D_{KL}\left(q_\phi(z_t|x_t) \,\middle\|\, p_\theta(z_t)\right) + \text{RMSE}(b_t, \hat{b}_t) \right]
\end{equation}
where $D_{KL}\left(q_\phi(z_t|x_t) \,\middle\|\,p_\theta(z_t)\right)$ denotes the KL divergence between the approximate posterior distribution $q_\phi(z_t|x_t)$, parameterized by $\phi$, and the prior distribution $p_\theta(z_t)$, parameterized by $\theta$. The root mean squared error (RMSE) between the actual trajectory points $b_t$ and the predicted trajectory points $\hat{b}_t$ is used to measure the reconstruction error over timesteps $n$ for $N$ training samples, as shown in \cref{eq11}. The parameter $\beta$ balances the influence of the KL divergence, allowing for control over the degree of regularization imposed on the latent space.
\begin{equation}
\label{eq11}
\begin{aligned}
RMSE=\sqrt{\frac{1}{N}\sum_{i=1}^{N}\sum_{j=1}^{n}\left \|b_i^{t+j}-\hat{b}_i^{t+j}  \right \|}
\end{aligned}
\end{equation}
\par
For the task of intention prediction loss $\mathcal{L_{\text{int}}}$ , the binary cross-entropy (BCE) is employed, given by \cref{eq12}. This choice is particularly well-suited for problems where the output can be categorized into one of two classes, such as predicting a pedestrian's intent to cross the street or not. The BCE loss function measures the divergence between the predicted probabilities, denoted as $\hat{I}$, and the actual ground truth labels, $I$, which are either 0 or 1.
\begin{equation}
\label{eq12}
\begin{aligned}
\mathcal{L_{\text{int}}}=-\frac{1}{N}\sum_{i=1}^{N}(I_ilog(\hat{I}_i)+(1-I_i)log(1-\hat{I}_i))
\end{aligned}
\end{equation}
The complete loss function for the proposed method is the sum of the trajectory bounding box prediction loss and the intention prediction loss:
\begin{equation}
\mathcal{L_{\text{total}}} = \lambda_{traj} \cdot \mathcal{L_{\text{traj}}} + \lambda_{int} \cdot \mathcal{L_{\text{int}}},
\end{equation}
where, $\lambda_{traj}$, and $\lambda_{int}$ are weighting parameters that balance the contributions of the trajectory bounding box prediction loss and the intention prediction loss, respectively. In our experiments, setting $\lambda_{traj}=1$ and $\lambda_{int}=1$, gives the better results.
This composite loss function is instrumental in concurrently optimizing pedestrian trajectory and intention prediction, which is vital for navigating dynamic and intricate environments.

\section{Experimentation and Results}
\subsection{Datasets}
The effectiveness of the method is evaluated using two specialized datasets for pedestrian behavior prediction in moving vehicles: the Pedestrian Intent Estimation (PIE) dataset \cite{rasouli2019pie} and the Joint Attention in Autonomous Driving (JAAD) dataset \cite{kotseruba2016joint}. The JAAD dataset consists of 346 high-resolution clips from 240 hours of driving footage, annotated at a 30Hz frame rate and focusing on 686 pedestrians with pedestrian behaviour annotations. These pedestrians are further divided into training, validation, and testing subsets, containing 188, 32, and 126 individuals, respectively. It provides comprehensive annotations, including pedestrian behaviors, poses, and scene-specific details like traffic signs. The PIE dataset is captured at a resolution of 1920 x 1080 pixels and a frame rate of 30 fps. It comprises over six hours of driving footage and features 1,842 annotated pedestrians. These are allocated across training, validation, and test sets, with counts of 880, 243, and 719 individuals, respectively. The PIE dataset includes not only annotations specific to pedestrians but also spatial metadata for other significant elements in the scene, such as traffic infrastructure and interacting vehicles. In both datasets, We have adopted the standard split as provided in the dataset.
\subsection{\label{sec:Training}Training Details}
The PTINet framework is trained on a GPU server utilizing the PyTorch library, and the network undergoes end-to-end training from scratch. An input horizon of $m=16$ timesteps, corresponding to 0.5 seconds, is considered, along with output horizons $n$ of 0.5, 1, and 1.5 seconds. Image data is resized to dimensions $[240, 420]$, and no other preprocessing or filtering is applied to the input images.  The optical flow is estimated using the PyTorch toolkit MMflow \cite{2021mmflow}, which encompasses a variety of state-of-the-art methodologies. 
After extensive experimentation and comparative analysis, we selected the  Recurrent All Pairs Field Transforms for Optical Flow (RAFT) \cite{teed2020raft} method due to its superior performance in capturing detailed motion patterns. The optical flow was computed between consecutive images. The optical flow was also resized to dimensions $[240, 420]$, and no other preprocessing was performed.
Pedestrian attributes, scene attributes, and pedestrian behavior data are employed in a categorical format.
Training optimization< is performed using the Adam optimizer, following the learning rate schedule as specified by $l_r=l_r^{int}\times (\frac{1-epoch}{max-epoch})^p$. Initial learning rate parameters $l_r^{int}$ are set at $0.0001$, with epsilon and weight decay values configured at $1^{-9}$ and $1^{-4}$, respectively. The power $p$  during the training phase is set at $0.9$. Training proceeds for $200$ epochs with a batch size of $4$.
% The input image sequence undergoes processing via a CLSTM Image module, as depicted in Figure\ref{framework}. This module comprises three Convolutional Long Short-Term Memory Network (C-LSTM) cells \cite{shi2015convolutional}. Each C-LSTM cell is followed by a max-pooling layer, with the exception of the last cell, which is succeeded by a fully connected layer instead. Convolutional kernels for each layer have a size of 5x5 and a stride of 2x2, with 32 filters. This module is especially well-suited for image sequences, as it is designed to learn both spatial and temporal dependencies concurrently. C-LSTM cell maintain a continuously updated hidden state as they process the input sequence, enabling them to model non-linear temporal transitions effectively. 
% \par 
% The optical flow is obtained using the consecutive frame 
 \begin{figure}
    \centering
    % Row 1
    \begin{subfigure}[b]{7cm}
        \includegraphics[width=7cm]{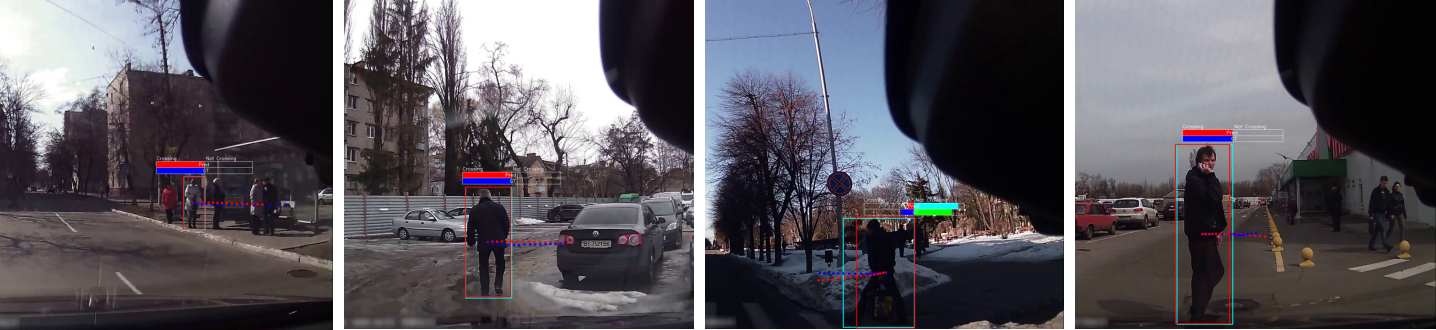}
        \caption{JAAD Dataset}
    \end{subfigure}
    \vspace{0.2cm}
    
    \begin{subfigure}[b]{7cm}
        \includegraphics[width=7cm]{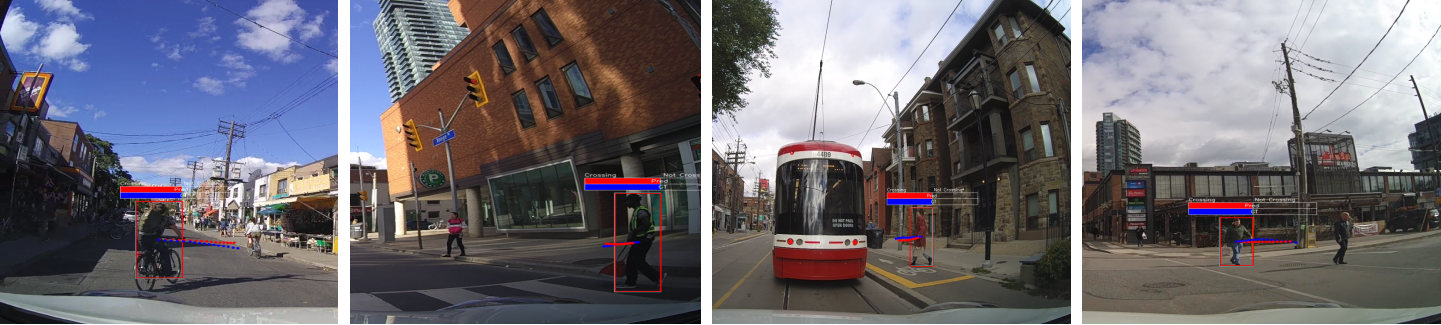}
        \caption{PIE Dataset}
    \end{subfigure}
    \vspace{0.2cm}
    \begin{subfigure}[b]{7cm}
        \includegraphics[width=7cm]{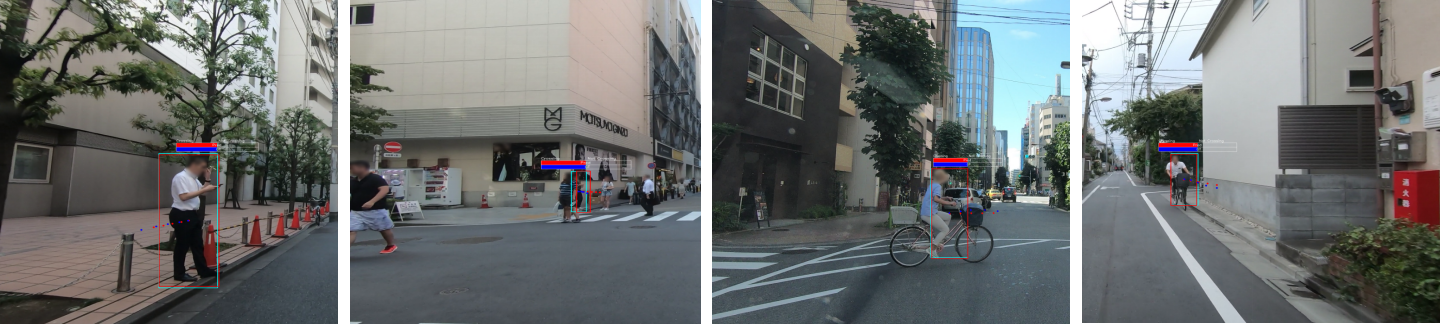}
        \caption{TITAN Dataset}
    \end{subfigure}
    \caption{The figure presents the qualitative results of the
proposed framework on the JAAD, PIE, and Titan datasets. Red
bounding boxes indicate predictions at the current timestamp,
while white bounding boxes represent ground truth values.
Dotted lines illustrate predicted trajectories over a 0.5s time
horizon, with blue indicating ground truth and red showing
predicted values. The bar graph displays the pedestrian’s
intentions, providing a comprehensive view of the model’s
performance.}
    \label{qualitative}
\end{figure}
\subsection{Evaluation Metrics}
To thoroughly evaluate the proposed methods, two distinct sets of metrics are applied, each tailored to specific aspects of the predictions. For trajectory prediction,\acfi{ade} and \acfi{fde} are employed, both calculated based on bounding box position. \ac{ade} measures the average Euclidean distance between the predicted and actual bounding box coordinates over a sequence of $n$ time steps. \ac{fde}, on the other hand, focuses only on the position in the final time step. All metrics are reported in pixels.
\par
For intention prediction, the F1 score and accuracy serve as the evaluation metrics, gauging the network's ability to identify pedestrian intentions correctly. The F1 score is calculated as the harmonic mean of precision and recall. The accuracy is a measure of the number of correct predictions out of the total number of instances. These metrics together offer a comprehensive evaluation of both the trajectory and intention prediction aspects of the network.
  \begin{table}[t]
    \centering
    \caption{Quantitative evaluation of the proposed method and state-of-the-art approaches on JAAD and PIE datasets for pedestrian intention prediction, focusing on F1-score and accuracy metrics.}
    \label{tab:intent-prediction}
    \resizebox{7cm}{!}{%
        \begin{tabular}{lcccc}
                    \toprule
        Methods & \multicolumn{2}{c}{JAAD } & \multicolumn{2}{c}{PIE }  \\
        \cmidrule(r){2-3} \cmidrule(lr){4-5} \\
          \textcolor{white}{Methods} & \multicolumn{1}{c}{F1-score } & \multicolumn{1}{c}{Accuracy }  & \multicolumn{1}{c}{F1-score } & \multicolumn{1}{c}{Accuracy } \\
         \midrule
            PCPA \cite{osman2023tamformer} & 0.67 & 0.56 & 0.77 & 0.86 \\
            R-LSTM \cite{osman2023tamformer} & 0.74 & 0.65 & 0.52 & 0.76 \\
            RU-LSTM \cite{osman2023tamformer} & 0.78 & 0.69 & 0.77 & 0.87 \\
            PIE-Intent \cite{rasouli2019pie} & - & - & 0.87 & 0.79 \\
            TAMformer \cite{osman2023tamformer} & 0.8 & 0.73 & 0.79 & 0.88 \\
            FF-STA \cite{yang2022predicting} & 0.74 & 0.62 & - & - \\
            BiPed \cite{rasouli2021bifold} & 0.6 & 0.83 & 0.85 & 0.91 \\
            PedFormer \cite{rasouli2023pedformer} & 0.54 & 0.93 & 0.87 & 0.93 \\
             \midrule
            % \textbf{Ours} & \textbf{0.91} & \textbf{0.95} & \textbf{0.96} & \textbf{0.98} \\
            \textbf{PTINet} & \textbf{0.92} & \textbf{0.96} & \textbf{0.965} & \textbf{0.98} \\ 
            % \textbf{OF} & \textbf{0.839} & \textbf{0.8} & \textbf{0.871} & \textbf{0.0.92} \\
            \hline
        \end{tabular}%
    }
\end{table}
\subsection{Results}

This section presents the evaluation results of the proposed context-aware multi-task learning framework on two publicly available datasets: JAAD and PIE. Figure \ref{qualitative} provides qualitative data to elucidate the performance of our proposed framework on the JAAD and PIE datasets. The bounding box in the figure indicates the current location of the pedestrian, while dotted lines represent predicted future trajectories. Bars in the figure indicate the pedestrian's intention, whether it is to cross or not to cross, over the considered time horizon.
The \cref{tab:trajectory-prediction}, shows the quantitative evaluation of the \ac{pti} with other state-of-art-methods. The results demonstrate that our framework outperforms state-of-the-art algorithms across varying time horizons ($0.5s$, $1s$, $1.5s$) for both \ac{ade} and \ac{fde} on the JAAD and PIE datasets. Specifically, our model \ac{pti} showcases the lowest \ac{ade} and \ac{fde} values in all examined time frames. In our evaluation, we compare our method with simple and complex models and those using multi-task learning frameworks and incorporate social attributes. In the JAAD dataset, the proposed method achieves better \ac{ade} and \ac{fde} scores at time horizon $(0.5s, 1s, 1.5s)$, outperforming the simple Linear model by a margin of $(95.1\%,97.4\%,98.2\%)$ for time horizons $(0.5s, 1s, 1.5s)$, respectively, on \ac{ade}. Similarly, in the case of \ac{fde}, the proposed method outperforms the Linear model by a margin of $98.4\%$ for a $1.5s$ time horizon, showing that the Linear model fails to capture the complexities of pedestrian behavior. In addition, methods such as SGNet and Bi-Trap-D, which mainly depend on trajectory data for their predictions and lack the incorporation of social behavior or pedestrian-centric features, outperform the Linear model. However, compared to our proposed method, they show higher values for \ac{ade} and \ac{fde}. Similarly, when compared with the proposed method, methods like PedFormer and BiPed, which use semantic segmentation, ego-motion, and trajectory data, have high \ac{ade} and \ac{fde} scores. Specifically, we outperform BiPed and PedFormer by improving \ac{ade} by roughly $20.44\%$ and $9.36\%$ and \ac{fde} by $15.93\%$ and $5.16\%$, respectively for $1s$ time horizon. 
In the PIE dataset, the proposed method also outperforms the state-of-the-art methods, as shown in \cref{tab:trajectory-prediction}. The proposed method obtains the \ac{ade} score of $4.26$, $9.49$, and $16.94$ for time horizons $(0.5s, 1s, 1.5s)$ respectively, whereas it achieves the \ac{fde} scores of $9.01$,$23.15$, and $49.025$ for the specified time horizons, outperforming the \ac{ade} and \ac{fde} scores of state-of-the-art methods.

\begin{figure*}[t]
    \centering
    % Row 1
    \begin{subfigure}[b]{4.5cm}
        \includegraphics[width=4.5cm]{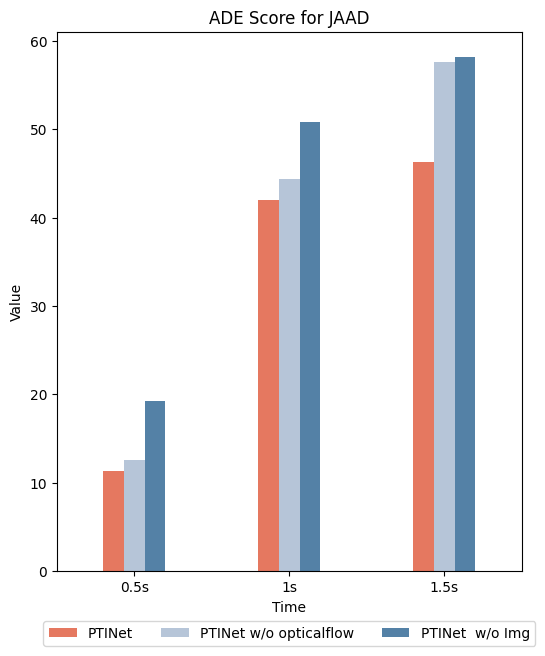}
        \caption{}
    \end{subfigure}
    % \hfill % this will add space between the subfigures
    \begin{subfigure}[b]{4.5cm}
        \includegraphics[width=4.5cm]{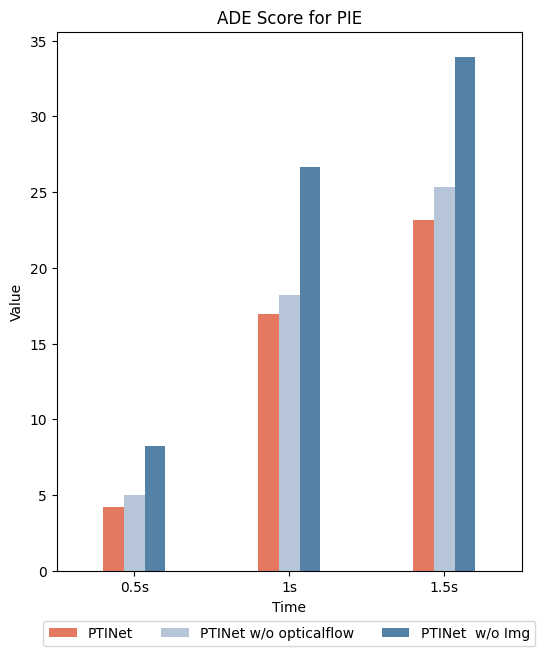}
        \caption{}
    \end{subfigure}
    % \hfill
    \begin{subfigure}[b]{4.5cm}
        \includegraphics[width=4.5cm]{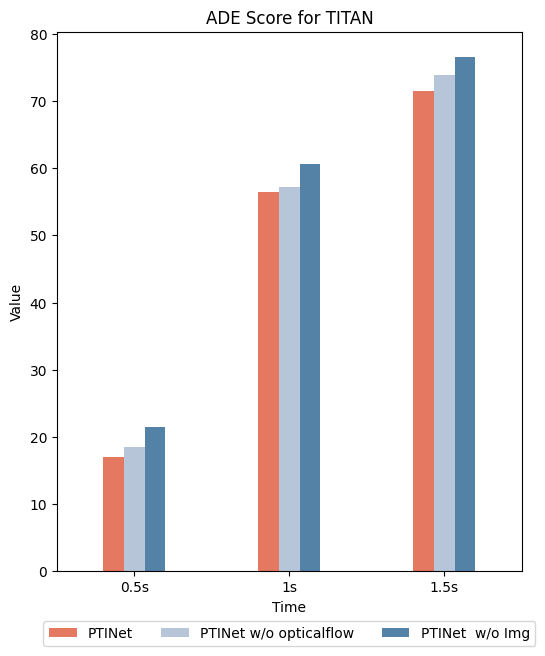}
        \caption{}
    \end{subfigure}

    % Add some space between the rows
    \vspace{0.2cm}
    
    % Row 2
    \begin{subfigure}[b]{4.5cm}
        \includegraphics[width=4.5cm]{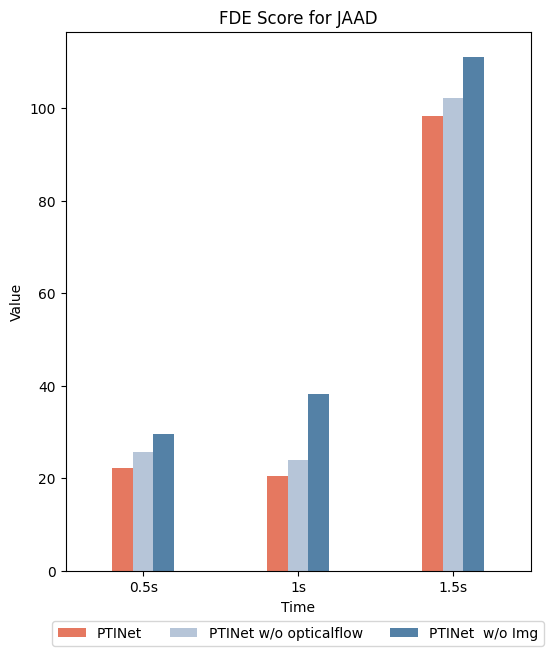}
        \caption{}
    \end{subfigure}
    % \hfill
    \begin{subfigure}[b]{4.5cm}
        \includegraphics[width=4.5cm]{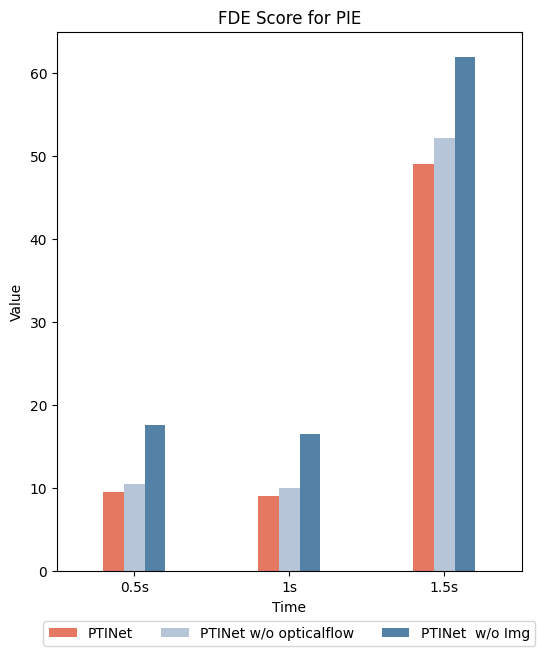}
        \caption{}
    \end{subfigure}
    % \hfill
    \begin{subfigure}[b]{4.5cm}
        \includegraphics[width=4.5cm]{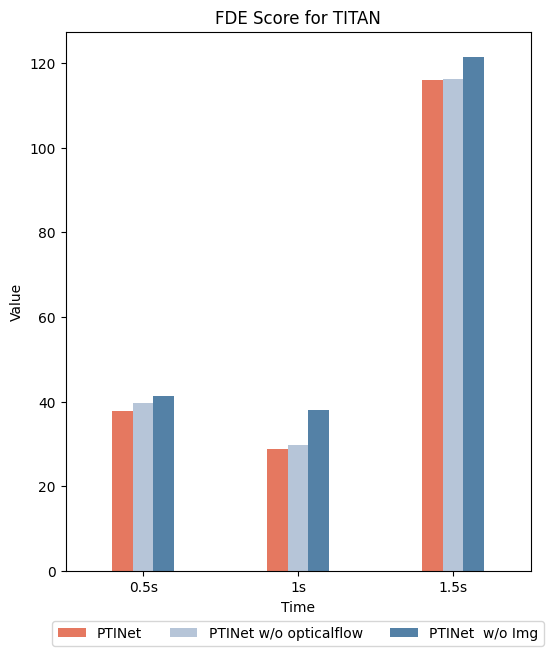}
        \caption{}
    \end{subfigure}
    \caption{Bar plots show the comparative performance of ADE and FDE scores for PTINet, PTINet without image data, and PTINet without optical flow on the JAAD, PIE, and TITAN datasets. }
    \label{bar-ade}
\end{figure*}

\par
The \cref{tab:intent-prediction} presents a quantitative evaluation of intention prediction algorithms on the JAAD and PIE datasets. The results indicate that our proposed framework achieves an F1-score of $0.92$ and an accuracy of $0.96$ on the JAAD dataset and an F1-score of $0.965$ and an accuracy of $0.98$ on the PIE dataset. In the JAAD dataset, TAMformer also shows promising results with an F1-score of $0.8$ and an accuracy of $0.73$. The model incorporates bounding boxes, poses, and local context and develops a transformer-based framework. Compared to TAMformer, our approach exhibits improvements of approximately $15\%$ in F1-score and $31.5\%$ in accuracy. PedFormer is another noteworthy algorithm. While it attains a high accuracy of $0.93$ on the JAAD dataset, its F1-score is $0.54$. This suggests that PedFormer may excel in some aspects of prediction, such as correctly identifying true positives and negatives, but may face challenges in minimizing false positives and negatives, which is a crucial factor for a balanced F1 score. For the PIE dataset, PedFormer and BiPed show robust performances. PedFormer has an F1-score of $0.87$ and an accuracy of $0.93$, while BiPed scores an F1-score of $0.85$ and an accuracy of $0.91$. Both algorithms benefit from multi-task learning and the mutual reinforcement of trajectory and intention prediction. 
\par
The results suggest that including pedestrian past trajectory, \ac{lcf} and \ac{gf} provides a holistic and more comprehensive understanding of pedestrian behavior. Additionally, the utilization of multi-task learning appears to offer mutual benefits for both trajectory and intention prediction tasks. 
\section{Ablation Study}
\begin{table}[t]
    \centering
    \caption{Quantitative evaluation of the proposed method and state-of-the-art approaches on TITAN datasets for pedestrian trajectory prediction, focusing on \ac{ade} and \ac{fde} metrics across time horizons of 0.5s, 1s, and 1.5s}
    \label{tab:titan}
    \resizebox{9cm}{!}{%
    \begin{tabular}{lcccccccc}
        \toprule
        Methods & \multicolumn{8}{c}{TITAN}  \\
        \cmidrule(r){2-9}   \\
          \textcolor{white}{Methods} & \multicolumn{3}{c}{ADE} & \multicolumn{3}{c}{FDE} & \multicolumn{1}{c}{F1-score } & \multicolumn{1}{c}{Accuracy} \\
        \cmidrule(r){2-4} \cmidrule(lr){5-7} 
        & 0.5s & 1.0s & 1.5s & 0.5s & 1.0s & 1.5s &\textcolor{white}{1.0s} & \textcolor{white}{1.0s} \\
        \midrule
        Bitrap \cite{rasouli2019pie} & 194 & 352 & 658 & - & - & 989 & - & -  \\
        ABC+\cite{rasouli2019pie} & 165 & 302 & 575 & - & - & 843 & - & -  \\
         \midrule
        % \textbf{Ours} & \textbf{18.4929} & \textbf{39.645} & \textbf{57.20} & \textbf{29.6652} & \textbf{73.9256} & \textbf{116.2543} & \textbf{0.95} & \textbf{0.97} \\
        \textbf{PTINet} & \textbf{16.977} & \textbf{37.786} & \textbf{56.412} & \textbf{28.79} & \textbf{71.596} & \textbf{115.924} & \textbf{0.96} & \textbf{0.975} \\ 
        % \textbf{OF} & \textbf{21.45} & \textbf{41.31} & \textbf{60.63} & \textbf{38.05} & \textbf{76.53} & \textbf{121.34} & \textbf{0.91} & \textbf{0.93} \\
        \bottomrule
    \end{tabular}%
    }
\end{table}
\subsection{Evaluation on TITAN Dataset }
To evaluate the generalization efficacy of our proposed approach, we conducted a case study on the TITAN dataset \cite{malla2020titan}. This dataset consists of $700$ video sequences captured via the front-view camera of a vehicle and provides bounding box annotations for $8,592$ distinct pedestrians, supplemented with contextual labels that characterize pedestrian attributes and behavioral patterns. It is imperative to note, that the dataset does not incorporate annotations for scene attributes, marking a limitation in the context of environmental feature analysis.
In this case, the \ac{lcf} consists of only pedestrian attributes and behavior. The dataset standard split as specified in \cite{malla2020titan} is used in our experimental design, allocating $400$ video sequences for training, $200$ for validation, and the remaining $100$ for testing purposes. For the extraction of optical flow, we employed the RAFT algorithm, as discussed before. We perform experimentation using the same hyper-parameters as discussed in \cref{sec:Training}.
The \cref{tab:titan} presents a quantitative analysis of pedestrian trajectory prediction on the TITAN dataset, comparing our method with the latest state-of-the-art approaches. Our method shows a significant improvement in accuracy. The \ac{ade} begins at $18.4929$ for a $0.5s$ prediction and increases to $57.20$ at $1.5s$. The \ac{fde} starts at $29.6652$ at $0.5s$ and goes up to $116.2543$ at $1.5s$. These results highlight our model's strong performance in making accurate predictions over time. Additionally, the F1-score and accuracy of our method are high, at $0.95$ and $0.97$, respectively, indicating the reliability of our model in predicting pedestrian trajectories and intentions accurately. The comparative analysis highlights the \ac{pti} performance in terms of trajectory accuracy over multiple time horizons and the reliable prediction of pedestrian intentions.

\subsection{Effect of Global Features on PTINet}

\begin{figure}
    \centering
    % Row 1
    \begin{subfigure}[b]{4cm}
        \includegraphics[width=4cm]{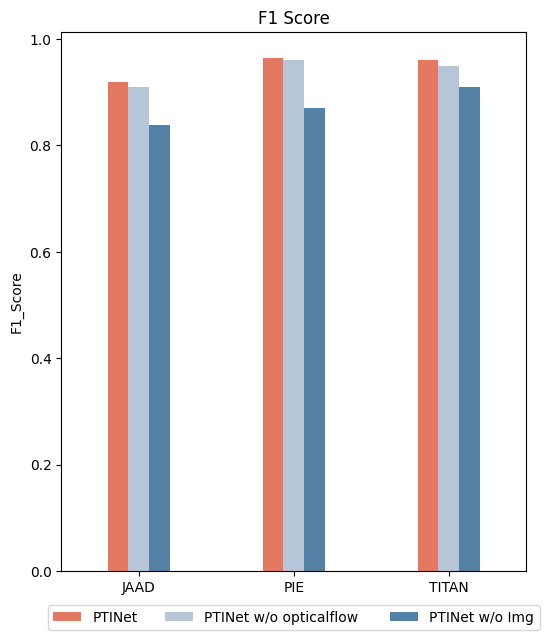}
        \caption{}
    \end{subfigure}
     % \hfill
    \begin{subfigure}[b]{4cm}
        \includegraphics[width=4cm]{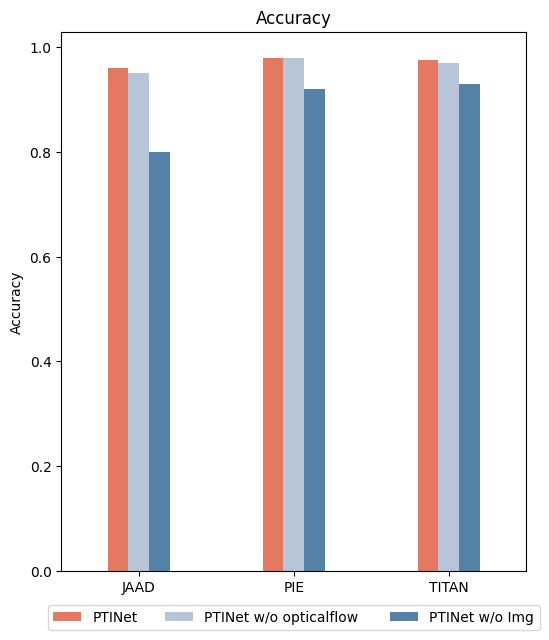}
        \caption{}
    \end{subfigure}
    % \hfill
    \caption{The bar plots show the comparison of F1 score and accuracy for intention prediction for PTINet, PTINet without image data, and PTINet without optical flow on the JAAD, PIE, and TITAN datasets.}
    \label{bar-f1}
\end{figure}

In our experiments, an ablation study was conducted to determine the influence of \ac{gf} on the performance of the  \ac{pti}. This evaluation involved analyzing the \ac{pti} performance in the absence of optical flow and, separately, without image data. The \cref{bar-ade} and \cref{bar-f1} illustrate the impact on \ac{ade} and \ac{fde} for trajectory prediction across the JAAD, PIE, and TITAN datasets and the F1-score and accuracy for intention prediction, respectively. The outcomes of this investigation show the significance of each feature towards enhancing prediction accuracy. The removal of optical flow resulted in a minor reduction in accuracy, highlighting its role in capturing the dynamic elements within scenes. While optical flow is beneficial for understanding movement patterns, its absence can be somewhat mitigated by other features within the model. On the other hand, the exclusion of image data led to a significant decline in the model’s performance. This is attributed to the critical role that image data plays in providing comprehensive contextual insights into the environment, which are indispensable for accurately forecasting pedestrian trajectories and intentions. Image data delivers crucial spatial and contextual cues necessary for decoding complex scenarios and anticipating future movements with precision, whereas optical flow contributes important but comparatively less essential information regarding temporal dynamics. This analysis emphasizes the integral role of both optical flow and image data in the predictive accuracy of \ac{pti}, with image data being especially crucial for maintaining model robustness and precision.

\subsection{Comparison with SOTA pedestrian trajectory prediction algorithms}
An ablation study was conducted using state-of-the-art pedestrian trajectory prediction algorithms on the JAAD and PIE datasets. After an exhaustive review of the literature, we opted for four algorithms: Social GAN, Trajectron++, MID, and Social Implicit. These algorithms were selected based on their proven performance on the benchmark ETH/UCY dataset. As discussed in the related work section, the selected algorithms are designed to predict pedestrian trajectories based solely on historical data. These algorithms have not been evaluated on datasets specifically intended for pedestrian intention prediction. To address this, evaluations were conducted on the JAAD and PIE datasets using publicly available source code. The \cref{sota} showcases the trajectory prediction results over a time horizon of $0.5s$ for both the JAAD and PIE datasets. Social Implicit (SI) and Social GAN emerged as top performers. Specifically, SI registered the lowest \ac{ade} and \ac{fde} values of $27.77$ and $50.03$, respectively, on the JAAD dataset, and $15.58$ and $29.94$ on the PIE dataset. Conversely, Trajectron++ and MID lagged in performance; Trajectron++ recorded an \ac{ade} of $206.66$ and an \ac{fde} of $245.51$ on JAAD, and an \ac{ade} of $115.32$ and an \ac{fde} of $180.47$ on PIE. Both SI and Social GAN model human behavior and incorporate important spatio-temporal variables, such as changes in human speed, the presence of nearby pedestrians, and average walking speed. These factors could significantly affect the prediction of pedestrian trajectories.
Conversely, Trajectron++ and MID primarily employ scene graph representations to model the spatio-temporal interactions of pedestrians with their environment. The lack of human behavioral considerations in the design of Trajectron++ and MID leads to their inferior performance on the JAAD and PIE datasets. However, it is worth noting that these results still lag behind those of Bifold and Pedformer, which utilize pedestrian-centric features to predict pedestrian trajectories.
\par
The observed performance patterns on the JAAD and PIE datasets highlight the pressing need for the development of algorithms that focus specifically on pedestrian behavior in urban scenarios. Such algorithms would be essential for enhancing the safety measures implemented in autonomous vehicles.

\begin{table}[t]
    \centering
    \caption{Comparison of pedestrian trajectory prediction results for state-of-the-art methods on JAAD and PIE datasets.}
    \label{sota}
    \resizebox{7cm}{!}{%
        \begin{tabular}{lcccc}
            \toprule
        Methods & \multicolumn{2}{c}{JAAD } & \multicolumn{2}{c}{PIE }  \\
        \cmidrule(r){2-3} \cmidrule(lr){4-5} \\
          \textcolor{white}{Methods} & \multicolumn{1}{c}{ADE@0.5 } & \multicolumn{1}{c}{FDE@0.5 }  & \multicolumn{1}{c}{ADE@0.5 } & \multicolumn{1}{c}{FDE@0.5 } \\
         \midrule
            Social GAN \cite{gupta2018social} & 29.25 & 52.86 & 17.25 & 31.15 \\
            Trajectron ++ \cite{salzmann2020trajectron++} & 206.66 & 245.51 & 115.32 & 180.47 \\
            Social Implicit \cite{mohamed2022social} & 27.77 & 50.03 & 15.58 & 29.94  \\
            MID \cite{gu2022stochastic} & 151.87 & 180.45 & 95.48 & 121.62  \\
            \hline
        \end{tabular}%
    }
\end{table}

\section{Conclusion}
% \comment{TK}{This is more of a summary than a conclusion. I believe it is necessary to dive a bit deeper into the findings of your work. Point out the shortcomings of the SOTA.}
This study presents a novel multi-task learning framework for predicting pedestrian trajectory and intention by considering their interdependent relationship. The framework incorporates diverse features that include pedestrian attributes, behaviors, scene characteristics, and global image and opticalflow features. The framework is thoroughly evaluated on JAAD and PIE datasets and achieved lower \ac{ade} and \ac{fde} scores compared to existing algorithms for trajectory prediction and higher F1 scores and accuracy for intention prediction. These results highlight the potential advantages of employing a context-aware, multi-task learning strategy for improved trajectory and intention prediction in complex urban environments and offer a holistic understanding of pedestrian behavior.

% or


% \appendices

% % you can choose not to have a title for an appendix
% % if you want by leaving the argument blank
% \section{}

% % use section* for acknowledgment
% \section*{Acknowledgment}


\ifCLASSOPTIONcaptionsoff
  \newpage
\fi



%\bibliographystyle{IEEEtran}


% \end{thebibliography}

\bibliographystyle{IEEEtran}
\bibliography{PTINet.bib}

% \begin{IEEEbiography}{Farzeen Munir}
% \end{IEEEbiography}

% \begin{IEEEbiographynophoto}{Farzeen Munir}
\begin{IEEEbiography}[{\includegraphics[width=1in,height=1.25in, clip,]{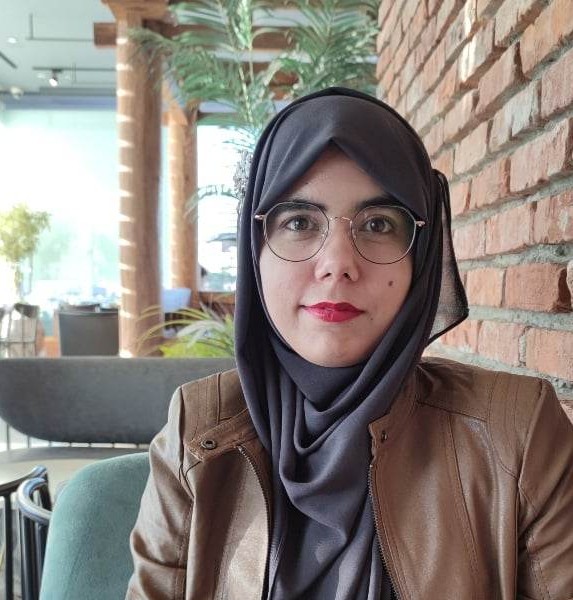}}]{Farzeen Munir} received her B.S degree in Electrical Engineering, and M.S. degrees in System Engineering from Pakistan Institute of Engineering and Applied Sciences, Pakistan, in 2013 and 2015, respectively. She completed her PhD from Gwangju Institute of Science and Technology, Korea in Electrical Engineering and Computer Science in 2022. Currently, she is working as a postdoctoral researcher at Aalto University, focusing on machine learning, deep neural networks, autonomous driving, and representation learning.
\end{IEEEbiography}
%\newpage
\begin{IEEEbiography}[{\includegraphics[width=1in,height=1.25in,clip,]{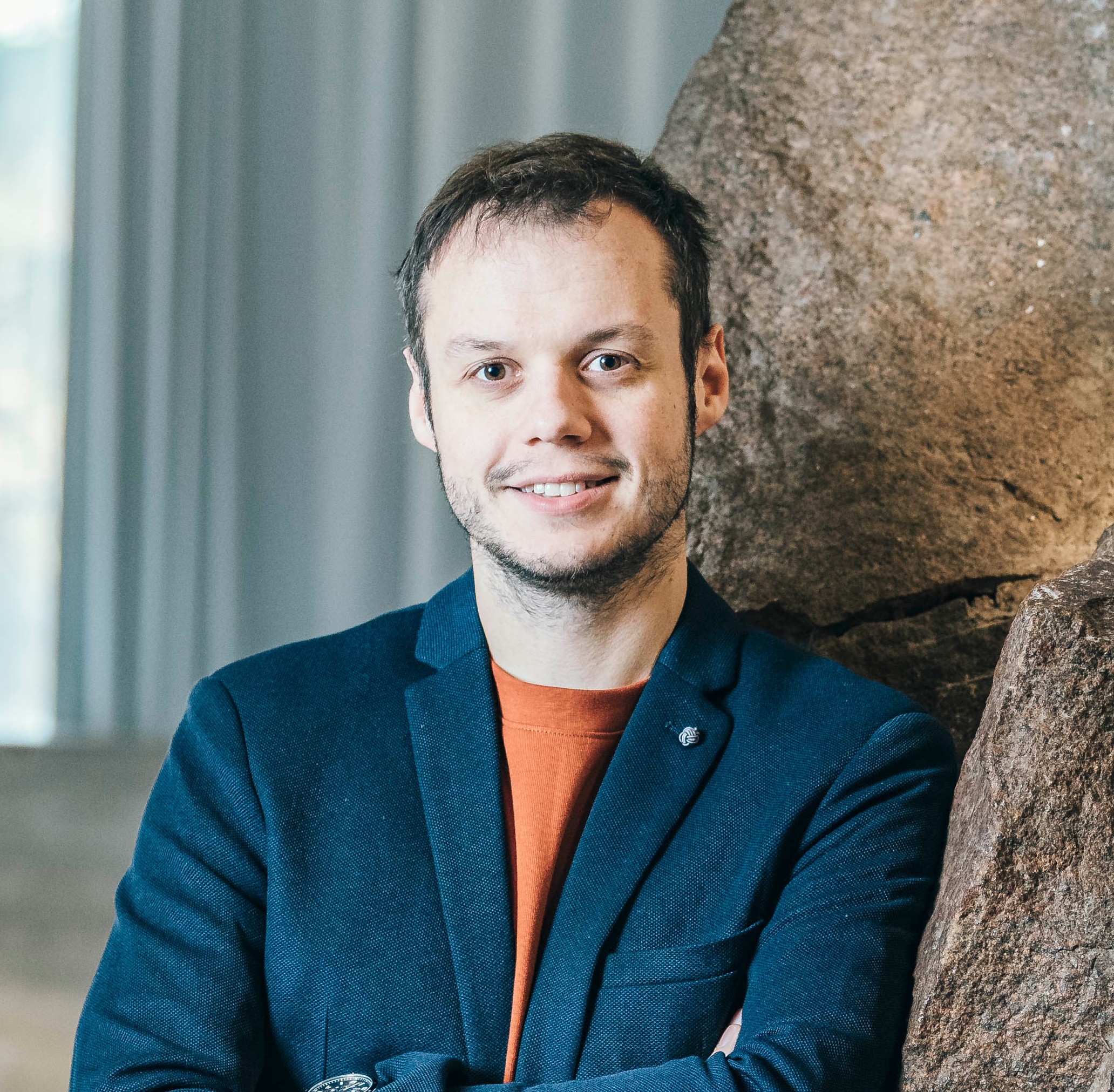}}]{Tomasz Piotr kucner} is an Assistant Professor at Aalto University, Finland in the field of robotics.  In 2018 he obtained a PhD in Computer Science from Örebro University, Sweden, Since 2021 he is leading the Mobile Robotics Group. In his research, he addresses problems related to the robust, reliable and legible operation of mobile robots in shared environments, including problems related to human motion prediction, building spatial models of human motion patterns, and human-aware motion planning. He is actively participating in the standardisation efforts as vice-chair of Vice-chair IEEE Robot 3D Map Data Representation Working Group of IEEE SA. 
\end{IEEEbiography}


%\vfill


\end{document}